# An Approach: Modality Reduction and Face-Sketch Recognition

**Sourav Pramanik**
*Assistant professor,*
*National Institute of Science and Technology,*
*Berhampur, India*
srv.pramanik03327@gmail.com

**Dr. Debotosh Bhattacharjee**
*Associate professor,*
*Jadavpur University*
*Kolkata, India*
debotosh@ieee.org

## Abstract

To recognize face sketch through face photo database is a challenging task for today's researchers. Because face photo images in training set and face sketch images in testing set have different modality. Difference between two face photos of difference person is smaller than the difference between same person in a face photo and face sketched. In this paper, for reduction of the modality between face photo and face sketch we first bring face photo and face sketch images in a new dimension using 2D Discrete Haar wavelet transform with scale 3 followed by a negative approach. After that, extract features from transformed images using Principal Component Analysis (PCA). Thereafter, we use SVM classifier and K-NN classifier for better classification. Our proposed method is experimentally verified by its robustness against faces that are captured in a good lighting condition and in a frontal pose. The experiment has been conducted with 100 male and female face images as training set and 100 male and female face sketch images as testing set collected from CUHK training and testing cropped photos and CUHK training and testing cropped sketches.

## 1. Introduction

Face recognition from still images and video sequences is emerging as an active research area over last 30 years with numerous commercial and law enforcement application. Face recognition systems can be used to allow access to an ATM machine or a personal computer, to control the entry of people into restricted areas, and to recognize people in specific areas (banks, stores), or in a specific database (police records) [1]. But in recent years face sketch recognition has become an active research area for engineers and scientist because in most cases the face image of a suspect is not available in the police records, than the best substitute is often a sketch drawing based on the recollection of an eyewitness is important, but sketch images are much different compared with original face photos in texture and shape. Therefore, it is difficult to match sketches and photos in the same modalities because they are belong in two different modalities [2]. Thus, automatically searching a potential suspect through a photo database using a sketch drawing becomes important. It can not only help police locate a group of potential suspects, but also help the witness and the artist interactively to modify the sketch during drawing based on similar photos retrieved [2], [3], [4], [5], [6], [7], [8], [9]. However, due to the great difference between sketches and photos and the unknown psychological mechanism of sketch generation, face sketch recognition is much harder than normal face recognition based on photo images.

To recognize a face sketch through photo database, it is necessary to reduce the modality difference between face photo and face sketch images. Every face recognition system need to perform mainly three subtasks: face detection, feature extraction and classification. But face sketch recognition system through face photo database need to perform mainly four subtasks: face detection, modality reduction, feature extraction and classification. In this paper, we have concentrated on three subtasks: modality reduction, feature extraction and classification. For modality reduction, it is necessary to bring all the face images in training set and all the face sketch images in testing set toward the new dimension. We have used efficiently computable 2D discrete Haar transform at scale 3 and a negative approach for modality reduction. We add an integer (I) to every image in testing set to get better result of modality reduction on new dimension. After that, we directly applied Principle Component Analysis on newly dimension images for dimensionality reduction or feature extraction. At last, SVM and K-NN classifiers have been used for classification or recognition task. The experiment has been conducted with 100 male and female face images as training set and 100 male and female face sketch images as testing set collected from CUHK training and testing cropped photos and CUHK training and testing cropped sketches.

The paper is organized as follows. Section 2, lists the survey on previous re-search that is most closely related to the present work and finds out problems. This is followed by a detailed description of the proposed system follows in section 3. Experimental results are presented in section 4. Finally, Section 5 ends the paper with several conclusions drawn from the design and the work with the proposed system.

## 2. Related work

There was only limited research work on face sketch recognition because this problem is more difficult than photo-based face recognition and no large face sketch database is available for experimental study. Methods directly using traditional photo-based face recognition techniques such as the eigenface methods [3] and the elastic graph matching methods [4] were tested on two very small sketch data sets with only 7 and 13 sketches, respectively. In [5] [6], a face sketch synthesis and recognition system using eigen transformation was proposed. In [7] proposed a nonlinear face sketch synthesis and recognition method. It followed the similar framework as in [5] [6]. The drawback of this approach is that the local patches are synthesized independently at a fixed scale and face structures in large scale, especially the face shape, cannot be well learned. In [8], [9] proposed an approach using an embedded hidden Markov model and a selective ensemble strategy to synthesize sketches from photos. The transformation was also applied to the whole face images and the hair region was excluded. In [2], proposed a face sketch synthesis and recognition approach based on local face structures at different scale using a Markov Random Fields model. But the drawback of this approach is that it requires a training set containing photo-sketch pairs. In [10], proposed an example-based face cartoon generation system. It was also limited to the line drawings and required the perfect match between photos and line drawings in shape. These systems relied on the extraction of face shape using face alignment algorithms such as Active Appearance Model (AAM) [11]. These line drawings are less expressive than the sketches with shading texture.

## 3. Proposed Approach

Recognition of face sketch images through face photo database is a difficult task compared with normal face recognition. Because face sketch and face photo images belongs to different modality. Thus, to recognize a face sketch we have first tried to reduce the modality difference between face photo and face sketch which is the most important part in face sketch recognition system. After that, features are extracted from newly dimension face photos and face sketches. After extraction of features, SVM and K-NN classifiers have used for classification. We have divided this section into four parts: preprocessing, modality reduction using wavelet transform followed by negative approach, dimensionality reduction or feature extraction using principle component analysis (PCA) and recognition task.

### 3.1 Preprocessing

Before extraction of the features from face photo and face sketch we have done some preprocessing task. In this step, photos are in color, first converted the RGB color to gray level and all the face photos and face sketches are resized with the dimension of $50 \times 65$ pixels. Fig.1 shows (a) color image, (b) corresponding gray image, (c) corresponding resized image. Fig.2 shows (a) a face sketch, (b) corresponding resized face sketch.

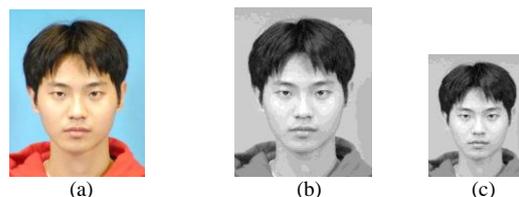

(a)　　　　　　(b)　　　　　　(c)

Fig.1 (a) original color face photo (b) corresponding gray face photo (c) corresponding resized face photo

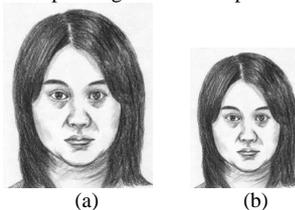

(a)　　　　(b)

Fig. 2 (a) original face sketch image (b) corresponding resized image

### 3.2 Modality Reduction using 2D Discrete Haar Transform followed by Negative Approach

Main problem to recognize a face sketch image through face photo database is the modality difference. For reduction of the modality between face sketch image and face photographs, we bring both types of images toward new dimension by using 2D Discrete Haar Transform followed by negative approach. In this work, first 2D Discrete Haar Transform (DHT) is applied on face photo and face sketch at scale 3 than we consider only diagonal coefficient. After that, negative image is calculated from diagonal coefficient images using the equation-1 and equation-2. At last, to improve the result of modality reduction on new dimension, an integer (I) is added to every sketch image. This section is divided into three parts: 2D Discrete Haar Theory, Haar Transform on face photo and face sketch, and negative approach.

#### 3.2.1 2D Discrete Haar Theory

Two dimensional discrete Haar Transform is just a composition of the one dimensional discrete Haar Transform twice in row direction and columns respectively. Let c be a M × L matrix. Also let $H^{row}$ and $G^{row}$ be the same matrix as H and G but operate on every row of c, and let $H^{col}$ and $G^{col}$ be the same matrix as H and G but operate on every column of c. Now for simplicity, assume c is always a square matrix that has $2^n$ rows [12]. Given J, N $\in$ **N** with J < N and a matrix $c_0 = \{c(m,n)\}_{n,m=0}^{2N-1}$. For $1 \leq j \leq J$, define the $2^{N-j} \times 2^{N-j}$ matrices $c_j, d_j^v, d_j^h, d_j^d$ by

$$c_j = H^{col} H^{row} c_{j-1},$$

$$d_j^v = G^{col} H^{row} c_{j-1},$$

$$d_j^h = H^{col} G^{row} c_{j-1},$$

$$d_j^d = G^{col} G^{row} c_{j-1},$$

Where $H = \frac{1}{\sqrt{2}}\begin{pmatrix} 1 & 1 & \cdots & 0 & 0 \\ \vdots & & \ddots & & \vdots \\ 0 & 0 & \cdots & 1 & 1 \end{pmatrix}$ and

$G = \frac{1}{\sqrt{2}}\begin{pmatrix} 1 & -1 & \cdots & 0 & 0 \\ \vdots & & \ddots & & \vdots \\ 0 & 0 & \cdots & 1 & -1 \end{pmatrix}$

### 3.2.2 Haar Transform on face photo and face sketch

In 2D Discrete Haar transform initially decompose an image into four components: approximated image (LL), the horizontal details information (LH), the vertical details information (HL) and the diagonal details information (HH). Among this four components diagonal details information of an image is the most balanced. In this work, 2D DHT is applied at scale 3 on face photo and face sketch images to get concise result and consider only diagonal coefficient of every image in training set and testing set which are our new dimension images. In mathematics an image can be represented by a function $f(x,y)$. In this work, let $f(n,m)$ and $g(n,m)$ is face photo and face sketch respectively. After applying 2D DHT at scale 3 and considering only diagonal coefficient, $f(n,m)$ and $g(n,m)$ can be expressed as $f'(n,m)$ and $g'(n,m)$ where,

$f'(n,m) = HH^3$ and $g'(n,m) = HH_3$.

Fig.3 and Fig.4 are showing the decomposition structure for face photo and face sketch image and Fig.5 and Fig.6 showing the coefficients at scale 3 for face photo and face sketch image.

| LL³ | LH³ | LH² | LH¹ |

| HL³ | HH³ | | |
| HL² | | HH² | |
| HL¹ | | | HH¹ |

Fig. 3 Decomposition structure at scale 3 for face photo

| LL₃ | LH₃ | LH₂ | LH₁ |
| HL₃ | HH₃ | | |
| HL₂ | | HH₂ | |
| HL₁ | | | HH₁ |

Fig. 4 Decomposition structure at scale 3 for face sketch image

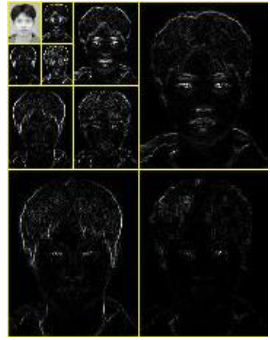

Fig. 5 Coefficients at scale 3 for face image

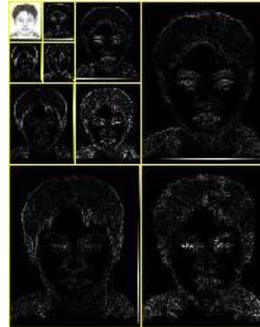

Fig. 6 Coefficients at scale 3 for face sketch image

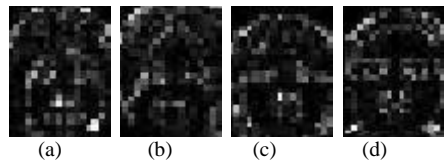

(a)  (b)  (c)  (d)
Fig. 7 (a)-(d) diagonal coefficients of face photos

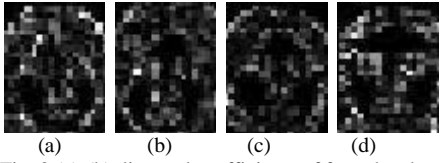

(a) (b) (c) (d)

Fig. 8 (a)-(b) diagonal coefficients of face sketches

### 3.2.3 Negative Approach

After considering the diagonal coefficient for face photo and face sketch images, to improve modality reduction negative image is calculated and an integer (I) is added with every face sketch image. Negative image is calculated for face photo and face sketch using the equation-1 and equation-2 and I is calculated using the equation-3.

$$\sim f'(n,m) = 255 - f'(n,m) \quad \ldots\ldots(1)$$

$$\sim g'(n,m) = 255 - g'(n,m) \quad \ldots\ldots(2)$$

$$I = |training\ set\ average\ value - testing\ set\ average\ value| \quad \ldots\ldots(3)$$

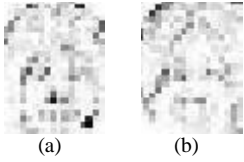

(a) (b)

Fig. 9 (a)-(b) negative image of two face photo

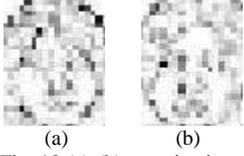

(a) (b)

Fig. 10 (a)-(b) negative image of two face sketch corresponding to the above two face photos

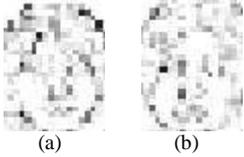

(a) (b)

Fig. 11 (a)-(b) negative image after adding I on two above face sketch

## 3.3 Dimensionality Reduction or Feature extraction Using Principle Component Analysis (PCA)

After bringing face photo and face sketch images toward new dimension, we have applied Principal Component Analysis (PCA) for feature extraction. We have applied PCA in a straight forward way. The steps for features extraction are as follows:

Step 1: Obtain M training images $I_1, I_2, \ldots, I_M$.

Step 2: Represent each image $I_i$ as a vector $\Gamma_i$. If images are $N \times M$ matrices, which can be represented as $NM \times 1$ dimensional vectors. $\Gamma_i = (x_1, x_2, \ldots, x_{NM})$

Step 3: Center the training images by subtracting the mean $m_i$ from each of the training image vector $\Gamma_i$.

$$\overline{\Gamma_i} = \Gamma_i - m_i$$

$$m_i = \frac{1}{MN} \sum_{j=1}^{MN} x_j$$

Step 4: Mean centered image vectors are now combined side-by-side, to create a data matrix Q of size $MN \times M$ (M is a number of training images).

$$Q = \overline{\Gamma_1}, \overline{\Gamma_2}, \ldots\ldots, \overline{\Gamma_M}$$

Step 5: Find the covariance matrix C:

$$C = QQ^T$$

Step 6: Calculate the eigen values and corresponding eigen vectors from the covariance matrix C by the equation

$$CV = \Lambda V$$

Where V is the set of eigen vectors associated with the eigen values $\Lambda$.

Step 7: Order the eigenvectors $v_i \in V$ according to their corresponding eigen values $\lambda_i \in \Lambda$ from high to low. Keep only the eigenvectors associated with non-zero eigen values.

Step 8: Each of the centered training images $\overline{\Gamma_i}$ is projected into the eigenspace. To project an image into the eigenspace, we have calculated the dot product $X_i$ of the centered image with each of the ordered eigenvector.

$$X_i = V^T . \overline{\Gamma_i}$$

Step 9: Each test image is also first mean centered by subtracting the mean image, and is then projected into the same eigenspace defined by V.

## 3.4 Recognition Task

After extraction of features from training and test images, we have used K-NN classifier with Mahalanobis distance measure and SVM classifier for classification or to recognize probe face sketch. The steps to recognize a query face sketch are as follows:

1. Face photo images and face sketch images are first brought to the new dimension using 2D Discrete Haar Transform followed by negative approach.
2. Then we have applied Principal Component Analysis (PCA) for dimensionality reduction or feature extraction from face photo and face sketch images.

3. After extraction of features from face photo and face sketch images, we have used K-NN classifier with Mahalanobis distance measure and SVM classifier with linear kernel function for classification or recognition of query face sketch image.

## 4. Experimental Results

In this phase, we provide our experimental results. We have divided our experimental result into two parts. The first one show the results for modality reduction and the second one shows the result for recognize faces corresponding to the query face sketch images.

### 4.1 Experimental Result for Modality Reduction

For modality reduction, we have used 2D Discrete Haar Transform followed by negative approach. We have tested this approach on 100 male and female face photos in training dataset and 100 male and female face sketch images in testing dataset. The training and testing images are collected from CUHK training and testing cropped photos and CUHK training and testing cropped sketches dataset. We have used Root Mean Square Error (RMSE) for performance evaluation of modality reduction between face photo and face sketch images as given by equation-4. In table-1, we compared the similarity measure between original face photo and face sketch and the corresponding new dimension face photo and face sketch images by giving the Root Mean square Error for each pair of face photo and face sketch shows in fig. 12 and fig. 13. Fig. 12 shows some pair of face photo and face sketch images collected from CUHK training and testing cropped photos and CUHK training and testing cropped sketches dataset. Fig. 13 shows corresponding new dimension images.

$$RMSE = \frac{\sqrt{\sum_{i=1}^{M}\sum_{j=1}^{N}[F(i,j)-S(i,j)]^2}}{M*N} \quad \ldots\ldots\ldots (4)$$

Where F is the face image and S is the sketch image.

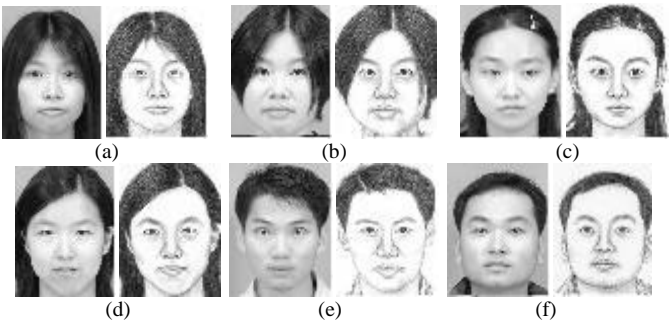

(a) (b) (c)

(d) (e) (f)

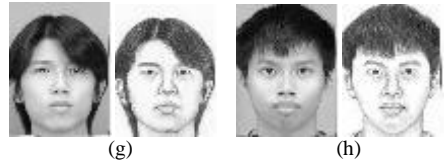

(g) (h)

Fig. 12 original pair of face photo and face sketch images

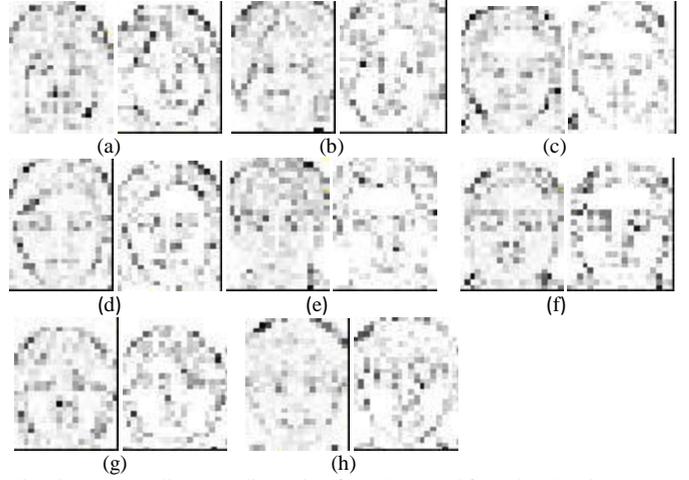

(a) (b) (c)

(d) (e) (f)

(g) (h)

Fig.13 corresponding new dimension face photo and face sketch pairs

TABLE-I. SIMILARITY MEASURE BETWEEN ORIGINAL PHOTO-SKETCH PAIRS AND NEW DIMENSION PHOTO-SKETCH PAIRS

| Pair no. | RMSE for original face photo and face sketch pairs | RMSE for corresponding new dimension pairs |
|---|---|---|
| a | 5.9161 | 2.7131 |
| b | 21.6102 | 12.8473 |
| c | 15.1658 | 8.1123 |
| d | 23.4307 | 9.5127 |
| e | 19.5959 | 7.3124 |
| f | 10.9211 | 4.1658 |
| g | 6.3246 | 1.8298 |
| h | 8.5440 | 2.4421 |

### 4.2 Experimental Result for Recognize a face sketch

In this section, we have given the recognition rate of our system and compared our method with two other methods. After extraction of features from new dimension images using Principal Component Analysis (PCA), we have applied two well known K-NN classifier and SVM classifier in a straight forward way for recognition of query face sketch image. Our method significantly good for 1st match in comparison with other two methods. Table-2 shows the matching percentage of the first five ranks of three methods. In table-2 we compare our proposed method with two methods: Nonlinear Face Sketch Recognition [7] and Sketch Transform Method [5]. The 1st match for Sketch Transform Method is no more than 80% and the 1st match for Nonlinear Face sketch Recognition is no more than 90%. Our method greatly improves the 1st match to 93%.

TABLE-II, CUMULATIVE MATCH SCORES OF FOUR METHODS

| Rank | 1 | 2 | 3 | 4 | 5 |
|---|---|---|---|---|---|
| Sketch Transform Method | 71 | 78 | 81 | 84 | 88 |
| Nonlinear face sketch recognition | 87.7 | 92.0 | 95.0 | 97.7 | 98.3 |
| New dimension +PCA+ K-NN | 91.4 | 93 | 93.5 | 94.1 | 95.4 |
| New dimension + PCA + SVM | 93 | 94.2 | 96.7 | 97 | 97.1 |

## 5. Conclusion

In this paper, we have proposed a novel method to recognize a face sketch, based on modality reduction. This is different and difficult than face photo recognition because faces are much different from sketches in terms of color, texture, and projection details of 3D faces in 2D images. To recognize a face sketch image through face photo database, we have first brought training and testing images toward the new dimension and then features are extracted from new dimension images using Principal Component Analysis (PCA). Finally, K-NN and SVM classifiers have been employed to recognize probe face- sketch through face photos database. To validate this new approach, the approach was tested using CUHK training and testing cropped photos and CUHK training and testing cropped sketches dataset. In our system, face-sketch is the input image and output is five face images which are best matching with input image. Curvelet based feature extraction may be employed in future for finding better features.